\documentclass[conference]{IEEEtran}
\IEEEoverridecommandlockouts
\usepackage{cite}
\usepackage{amsmath,amssymb,amsfonts}
\usepackage{algorithmic}
\usepackage{graphicx}
\usepackage{textcomp}
\usepackage{xcolor}
\usepackage{stfloats}
\usepackage{booktabs}
\usepackage[T1]{fontenc}
\usepackage[colorlinks,
linkcolor=red,
anchorcolor=blue,
citecolor=green
]{hyperref}
\def\BibTeX{{\rm B\kern-.05em{\sc i\kern-.025em b}\kern-.08em
    T\kern-.1667em\lower.7ex\hbox{E}\kern-.125emX}}
\begin{document}

\title{Learning Multiscale Correlations for Human Motion Prediction}
\author{\IEEEauthorblockN{Honghong Zhou\textsuperscript{1},
		Caili Guo\textsuperscript{1,2}\textsuperscript{*}, Hao Zhang\textsuperscript{1} and Yanjun Wang\textsuperscript{3}}
	\thanks{\textsuperscript{*}Corresponding author.}
	\IEEEauthorblockA{\textsuperscript{1}Beijing Key Laboratory of Network System Architecture and Convergence,\\
	School of Information and Communication Engineering,\\
	Beijing University of Posts and Telecommunications, Beijing, China\\
	\textsuperscript{2}Beijing Laboratory of Advanced Information Networks, Beijing, China\\
	\textsuperscript{3}China Telecom Dict Application Capability Center, China\\
	\{zhouhonghong, guocaili, zhanghao0215\}@bupt.edu.cn, wangyanjun@chinatelecom.cn} 
	}

\maketitle

\begin{abstract}
In spite of the great progress in human motion prediction, it is still a challenging task to predict those aperiodic and complicated motions. We believe that capturing the correlations among human body components is the key to understand the human motion. In this paper, we propose a novel multiscale graph convolution network (MGCN) to address this problem. Firstly, we design an adaptive multiscale interactional encoding module (MIEM) which is composed of two sub modules: scale transformation module (STM) and scale interaction module (SIM) to learn the human body correlations. Secondly, we apply a coarse-to-fine decoding strategy to decode the motions sequentially. We evaluate our approach on two standard benchmark datasets for human motion prediction: Human3.6M and CMU motion capture dataset.  The experiments show that the proposed approach achieves the state-of-the-art performance for both short-term and long-term prediction especially in those complicated action category. We make codes publicly available at \url{https://github.com/zhouhongh/MGCN}.

\end{abstract}

\begin{IEEEkeywords}
Human motion prediction, multiscale, graph convolution network, DCT
\end{IEEEkeywords}

\section{Introduction}

Human motion prediction aims to use the 3D skeleton data to predict a sequence of future human motions based on observed motion frames. It plays a significant role in robotics, computer graphics, healthcare and public safety\cite{b1,b2,b3} such as human robot interaction\cite{b4}, autonomous driving\cite{b5}, pedestrian tracking\cite{b6} etc. 

Traditionally, Hidden Markov Model\cite{b10} and Gaussian Process latent variable models\cite{b11} is used to predict human motions, but limited to simple actions such as walking and golf swing. More complicated actions are typically tackled using deep networks including the recurrent neural networks (RNNs) \cite{b13,b14,b15,b16,b17,b18,b19,b20} and feed-forward networks (FNNs)\cite{b21,b22,b23,b24,b25,b26,b27,b28}. Simply apply RNN without the modeling of the body structure especially the correlations among human body suffered bad results \cite{b13}. Jain proposes to use the Structural RNN to model relationship among the spine and  limbs which achieves good performance \cite{b19}. More and more people aware the importance of exploring the human body correlations from the body structure and proposed many RNN-based methods \cite{b15, b16,b17,b18}. Due to the RNN's weakness on capturing long-term temporal dependencies and the disadvantage of  error accumulation, the feed-forward networks have attracted more and more attentions.  As in \cite{b24,b25,b26,b27}, the Discrete Cosine Transformation (DCT) becomes the popular temporal encoding strategy, which allows the network to concentrate on extracting  the spatial correlations. Although achieved great progress, they all model the human motion in one single scale and the prediction on more complicated and aperiodic actions such as greeting and directing traffic is also a challenging task especially in the long-term scenario.

\begin{figure}[htp]
	\centerline{\includegraphics[width=8cm]{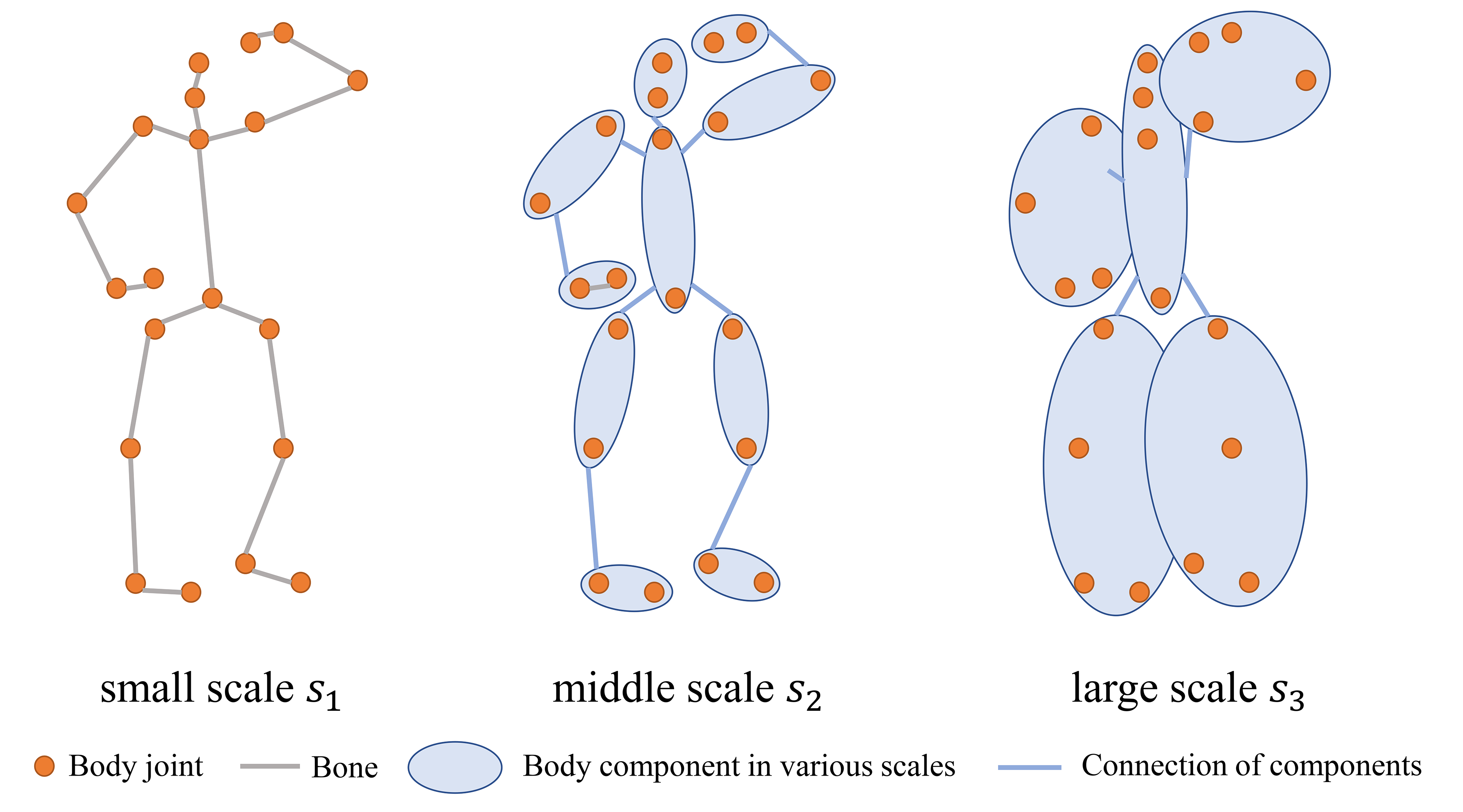}}
	\caption{Three body scales on Human 3.6M. In $s_{1}$, we consider 20 joints with non-zero angles\cite{b15}; In $s_{2}$ and $s_{3}$, we consider 10 and 5 components, respectively.
		\label{fig1}}
\end{figure}
\begin{figure*}[ht]
	\centerline{\includegraphics[width=16cm]{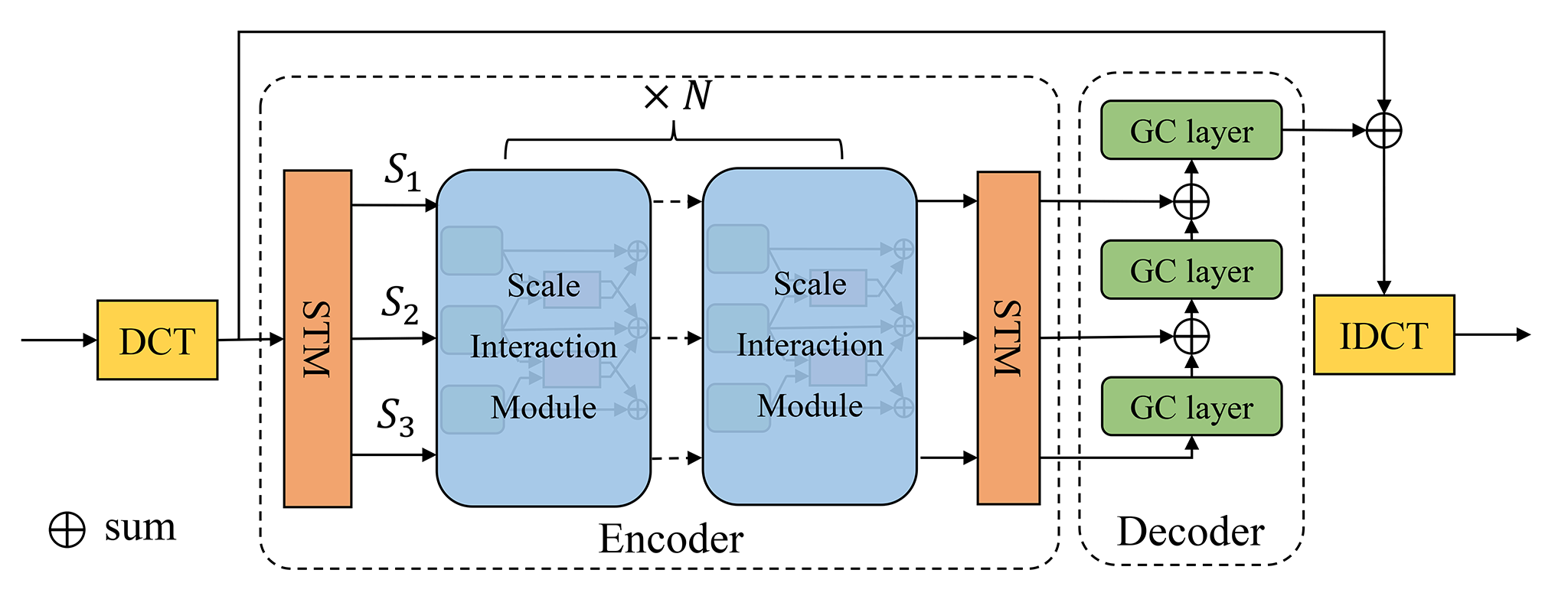}}
	\caption{The architecture of MGCN.
		\label{fig2}}
\end{figure*}

 We keep under observation on motion rules of the real person, found that in some actions like running, the co-movement mainly exists among the big limbs, while in some other actions like smoking, the small movement of the wrist or elbow could lead to really different future poses. This scalable attention of human motions inspires us to capture the correlations of human body in a multiscale way. In this paper, based on the multiscale graph proposed in \cite{b28} shown in Fig.~\ref{fig1}, we further exploit the multiscale modeling method and propose the multiscale graph convolution network (MGCN) as in Fig.~\ref{fig2}, which achieve much better performance than \cite{b28}.
 
In summary, our contributions are twofold: 
\begin{itemize}
	\item We make a comprehensive study on human motion prediction and propose an encoder-decoder framework called MGCN to deeply exploit the correlation of human body joints and components in the multiscale graphs, which is action-agnostic and end-to-end.
	\item We verify the effectiveness of MGCN on two benchmark datasets. Especially, our approach outperforms the SOTA method\cite{b26} 5 to 7 millimeters for the long-term prediction, which is a big lift.
\end{itemize}

\section{Related work}
In this section, we introduce the related works from the perspective of scales, including the single-scale and multiscale methods.
\subsection{Single-scale methods}

\subsubsection{RNN based methods}
The LSTM-3LR and ERD proposed by Fragkiadaki et al. \cite{b13}  model the human motion based on concatenated LSTM units. S-RNN \cite{b19} models the main five components of human body by five RNNs and exchange information among them. After the S-RNN \cite{b19}, more attention is payed on the capturing of the spatial dependencies especially the correlations of human body components. Liu et al. \cite{b15} design a hierarchical RNN which allows the information flows along adjacent joints and frames simultaneously. The evolution of these RNN based methods indicates the importance of modeling the human body correlations. While limited to RNN's drawbacks on error accumulation and long-term dependencies  extracting, the FNN based methods become more popular these years.

\subsubsection{FNN based methods}
Li et al.\cite{b22} model the human motions with the feed-forward convolutional network and gain lower prediction errors than the existing RNN methods. Influenced by the interests in signal processing, Mao et al.\cite{b24} propose to learn the body joints trajectory dependencies with DCT and apply the fully-connected GCN to explore the correlations among body joints, which brings impressive progress. Inspired by Mao et al. \cite{b24},  Lebailly et al. \cite{b25} introduce additional temporal inception module to achieve better temporal encoding. Cai et al.\cite{b25} replace the GCN with the transformer network and achieves the state-of-the-art performance. The success of GCNs and transformer networks also shows the importance of modeling the human body correlations.

\subsection{Multiscale methods}
The multiscale modeling strategy is widely used in machine learning, such as object detection\cite{b31,b32}, and NLP\cite{b33,b34}. It has a huge advantage over solving problems  which have important features at multiple scales of time and/or space. On the human motion prediction task, Li  et al. \cite{b28} propose to capture the human body correlations by the multiscale graph based on the backbone ST-GCN \cite{b35}. They generate human body graphs for three scales by mean-pooling, use the ST-GCN to encode information in each scale and design the cross-scale fusing blocks to fuse features with the adjacent scales. However, the mean-pooling strategy for generating multiscale graphs will cause information loss, and the ST-GCN mixes the temporal and spatial information, which is not benefit to the sequential human motion prediction task \cite{b36}. 

From the previous works we can draw a concluision that the key to predict motions accuratelly is to properly model the human body corelations and the multiscale methods have great potential on human motion prediction. The mentioned background shows the big value of our work and we introduce it detailly in the next section.

\section{Methodology}

 On this task, we assume to be given a history motion sequence $\mathbf{X}_{1: N}=\left[\mathbf{x}_{1}, \mathbf{x}_{2}, \mathbf{x}_{3}, \cdots, \mathbf{x}_{N}\right]$ consisting of N consecutive human poses, where $\mathbf{x}_{i} \in \mathbb{R}^{K}$, with $\mathbf{K}$  data dimensions describing pose at each time step. And our goal is to predict the future poses $\mathbf{X}_{N+1: N+T}$ for the future $T$ time steps. Before sending the input data to the MIEM, following \cite{b24}, we replicate the last pose $ \mathbf{x}_{N}$, $T$ times to generate a new sequence of length $N + T$ : $\mathbf{X^{\prime}}_{1: N+T}=\left[\mathbf{x}_{1}, \mathbf{x}_{2}, \cdots, \mathbf{x}_{N-1}, \mathbf{x}_{N},  \mathbf{x}_{N}, \cdots, \mathbf{x}_{N}\right]$ and compute the DCT coefficients of length $D$ for the new sequence as: 
\begin{equation}
\mathbf{F} = f_{DCT}(\mathbf{X^{\prime}}_{1: N+T})\in \mathbb{R}^{K \times D}
\label{eq0}
\end{equation}
where $\mathbf{F}$ is the DCT coefficients, and $f_{DCT}$ is the Discrete Cosine Transformation.

We make efforts to capture the human body correlations  with the proposed MGCN whose architecture is shown in Fig. ~\ref{fig2}. We use those replicated DCT coefficients to predict the real ones and finally make IDCT to obtain the human motion frames on Euler angle representation or 3D coordinates. 

\subsection{Encoder}
The MIEM plays the role of encoder. The MIEM includes two types of sub modules: 1) STM, which aggregates joints to components or convert components back into joints, and 2) SIM, which extract the human body correlations in and across three scales.
\begin{figure}[htp]
	\centerline{\includegraphics[width=8cm]{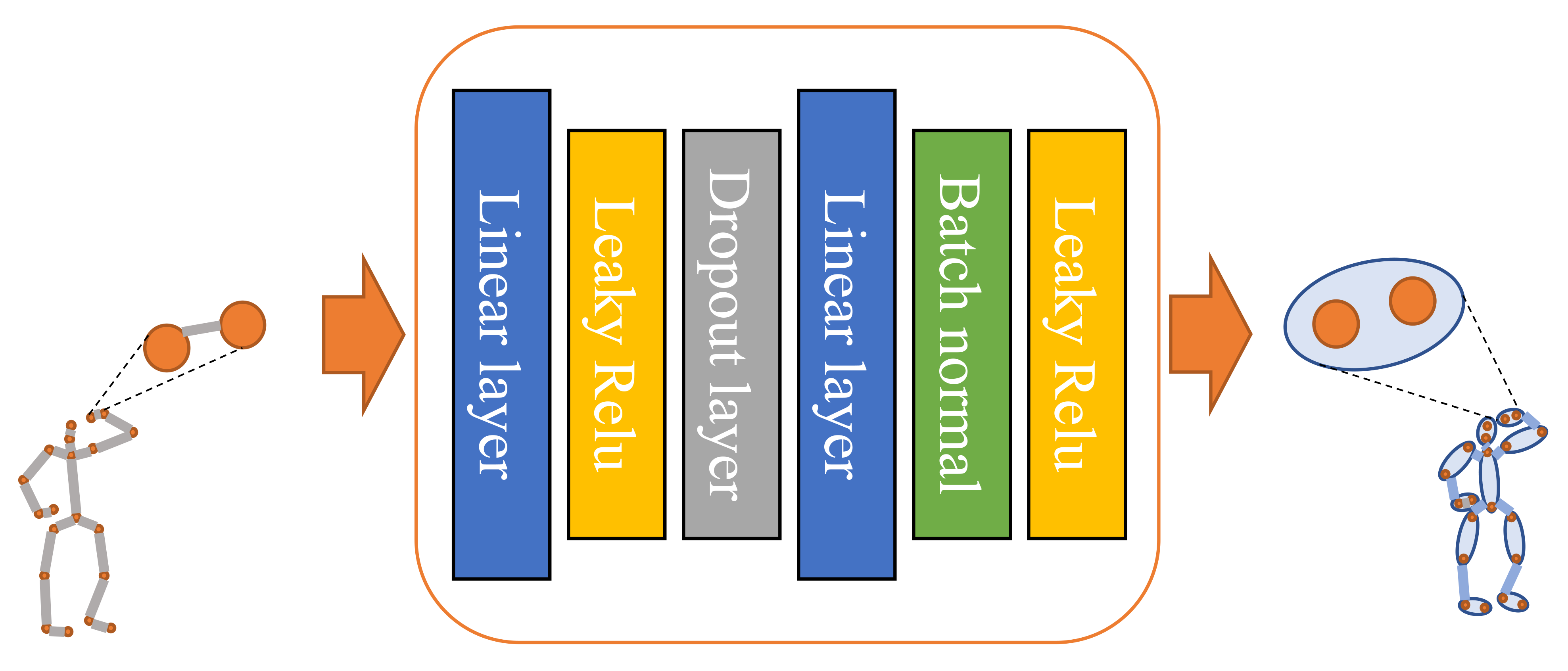}}
	\caption{STM core network. This figure shows how the STM transform 2 joints of $s_{1}$ into a component of $s_{2}$. Actually it is a two-layer MLP, for each component of $s_{2}$, we apply a MLP like this.
		\label{fig3}}
\end{figure}

\subsubsection{Scale transformation module (STM)}
We design two STMs for our encoder according to Fig.~\ref{fig2}, one is used to aggregate body joints into body components (before SIM) and the other is used to convert the aggregated body components back to body joints (afer SIM). As shown in Fig.~\ref{fig3}, the body joints at $s_{1}$ that belong to the same component are transformed to a new graph node at $s_{2}$ or $s_{3}$. For example, we can aggregate the ``right shoulder" joint and the ``right elbow" joint to constitute the ``right up arm" component. We apply a MLP for each component, since there are 10 body components at $s_{2}$ and 5 at $s_{3}$, we apply totally 15 MLPs to realize the graphs' scale transformation. Taking $s_1$ and $s_2$ as an example, the STM can be described by \eqref{eq1}:
\begin{equation}
\mathbf{F}_{k}^{s_{2}} = f_{1}(\mathbf{F}_{i:j}^{s_{1}})
\label{eq1}
\end{equation}
where $\mathbf{F}_{k}^{s_2}$ indicates the features of the $k^{th}$ node at $s_2$, $\mathbf{F}_{i:j}^{s_1}$ is the features concatenated between the $i^{th}$ and the $j^{th}$ node at $s_1$, $f_1$ is the two-layer MLP as Fig.~\ref{fig3}. The mapping between $k$ and $i,j$ is defined in advance according to the human body structure.

When it comes to transform $s_2$ and $s_3$ back to the size of $s_1$, the operation is similar. We just swap input dimension of MLPs with the output dimension. At $s_2$, this process can be depicted by \eqref{eq2}:
\begin{equation}
\mathbf{F^{\prime}}_{i:j}^{s_2} = f_{2}(\mathbf{F}_{k}^{s_2})
\label{eq2}
\end{equation}
where ${F^{\prime}}_{i:j}^{s_2}$ is the features at $s_2$ but has the same number of node as that at $s_1$. Noticed that  ${F^{\prime}}_{i:j}^{s_2}$ still indicates the features of scale 2, we remain the $s_2$ superscript.

\subsubsection{Scale interactional module (SIM)}
\begin{figure}[htp]
	\centerline{\includegraphics[width=8cm]{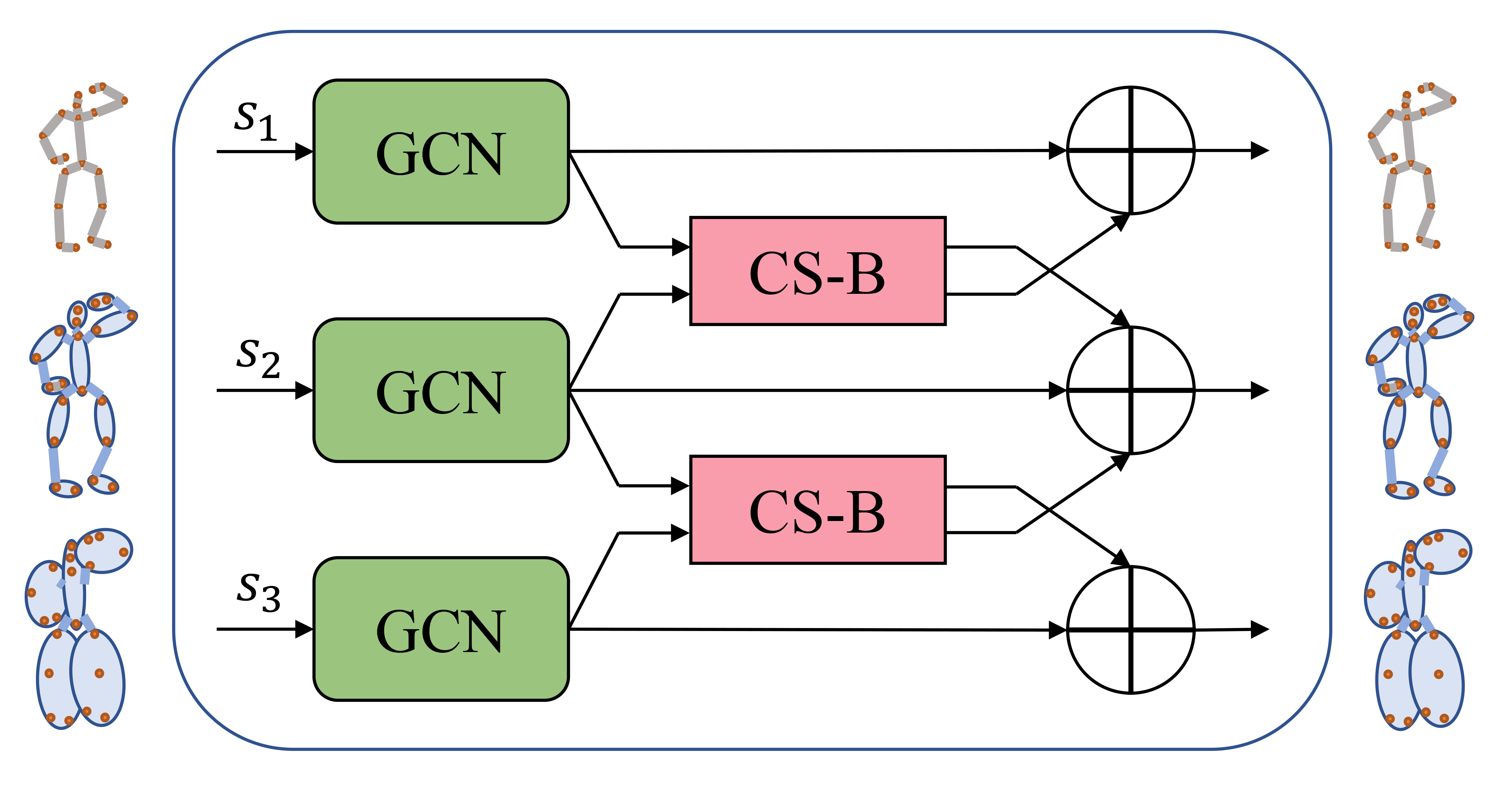}}
	\caption{The SIM network. The features of each scale get through a six-layer GCN, and relies on cross-scale blocks (CS-Bs) to explore the human correlations cross scales.
		\label{fig4}}
\end{figure}

In order to exploit the human body correlations more adequately, we design the SIM as Fig.~\ref{fig4}, and cascade it $N$ times to improve the feature extraction ability. the SIM is composed of two parts: the GCNs, which extract dynamic features in single scale, and the CS-Bs, which introduce the additional supervisory information from the adjacent scales.  As we all know that if we want to recognize someone's action, we need not only the cooperation among the large limbs such as arms and legs but also some subtle movements like the rotation of wrists, so the information-interacting strategy conforms our cognitive rules.

Following the notation of \cite{b24}, we model the skeleton as a fully-connected graph of $K$ nodes, represented by the trainable weighted adjacency matrix $\mathbf{A}^{K \times K}$. And the GCN is stacked by several GC layers, each performing the operation:
\begin{equation}
\mathbf{F}^{(p+1)}=\sigma\left(\mathbf{A}^{(p)} \mathbf{F}^{(p)} \mathbf{W}^{(p)}\right)
\label{eq3}
\end{equation}
where $\mathbf{W}^{(p)}$ is the set of trainable weights of layer $p$, $\mathbf{A}^{(p)}$ is the learnable adjacency matrix of layer $p$, $\mathbf{F}^{(p)}$ indicates the input of layer $p$ while $\mathbf{F}^{(p+1)}$ the input of layer $p+1$ (and the output of layer $p$), $\sigma(\cdot)$ is an activation function such as $tanh(\cdot)$.

\begin{figure}[htp]
	\centerline{\includegraphics[width=8cm]{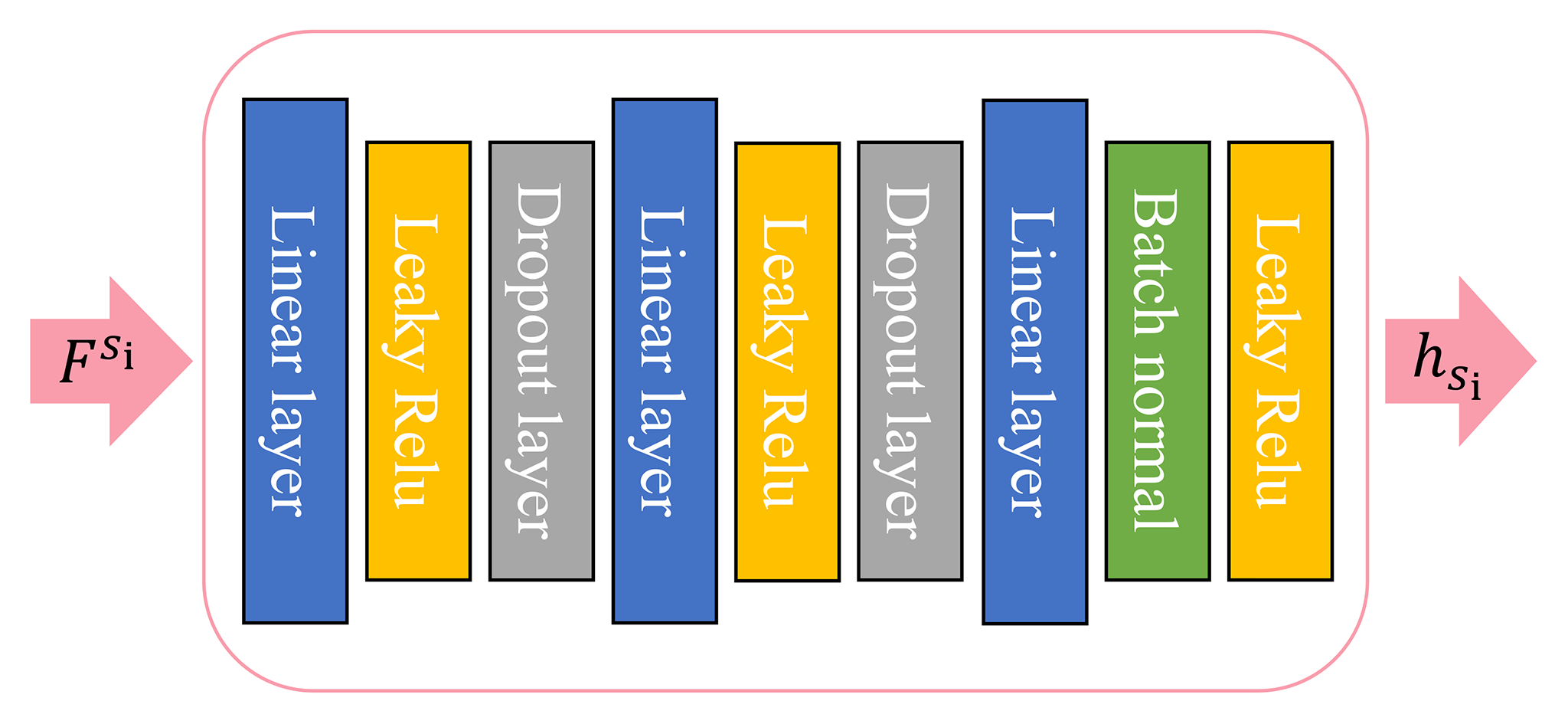}}
	\caption{The three-layer MLP that designed in the CS-B.
		\label{cs-b}}
\end{figure}

After the encoding of GCNs, the features interact cross scales by the cross-scale block (CS-B). Different from the existing complicated operations in \cite{b28}, we design three-layer MLPs as Fig.~\ref{cs-b} to get attention matrix $\mathbf{A}_{s_{i}s_{j}}$ for two adjacent scales, which can speed up the calculation and be easy to train. The process to generate $\mathbf{A}_{s_{i}s_{j}}$ can be described by:
\begin{equation}
h_{s_1}= f_{MLP_1}(F^{s_1}) \in \mathbb{R}^{K_{s_{1}} \times D_{h}}
\label{eq4}
\end{equation}
\begin{equation}
h_{s_2}= f_{MLP_2}(F^{s_2}) \in \mathbb{R}^{K_{s_{2}} \times D_{h}}
\label{eq5}
\end{equation}
\begin{equation}
\mathbf{A}_{s_{2}s_{1}} = softmax(h_{s_1}^\mathrm{T} h_{s_2}) \in \left[0,1\right]^{K_{s_{2}} \times K_{s_{2}}}
\label{eq6}
\end{equation}
where $ f_{MLP_1}$ and $ f_{MLP_2}$ denote MLPs as Fig.~\ref{cs-b}, $K_{s_{i}}$ means the number of nodes at $s_{i}$.  $\mathbf{A}_{s_{2}s_{1}}$ is the attention matrix from $s_2$ to $s_1$. Benefit from the attention matrixs $\mathbf{A}_{s_{2}s_{1}}$, we can now adaptively explore the cross-scale human body correlations in a distinct way.
We next introduce the supervisory information of the adjacent scale $s_2$ to $s_1$ with $\mathbf{A}_{s_{2}s_{1}}$, the feature at $s_1$ is updated as:
\begin{equation}
\mathbf{F}^{s_{1}} \leftarrow \mathbf{A}_{s_{2}s_{1}} \mathbf{F}^{s_{2}} +\mathbf{F}^{s_{1}} \in \mathbb{R}^{K_{s_{1}} \times D^{\prime}}
\label{eq7}
\end{equation}

\subsection{Decoder}
We apply the coarse-to-fine strategy to decode the features of three layers. The motivation of the decoder is that the larger scale can provide information of global motion evolution, which can indicate the approximate moving direction, speed and action category. The smaller scale then is able to predict the precise joint location with the global supervisory information.
As in Fig.~\ref{fig2}, the motion features at $s_{3}$ was sent to the bottom GC layer to predict the coarse future motion. Afterwards, we sum the output of the bottom GC layer and the feature at $s_{2}$ (middle scale) as the input of the middle GC layer. Similarly, the output of the top GC layer predicts the finest future motion. Mathematically, given the transformed features at $s_2$ and $s_3$, $\mathbf{F^{\prime}}^{s_2}$ and $\mathbf{F^{\prime}}^{s_3}$, and the feature at $s_1$, $\mathbf{F}^{s_1}$, the predicted DCT coefficients $\mathbf{F}_p$ are :
\begin{equation}
\mathbf{F}_p = f_{s_{1}}(f_{s_{2}}(f_{s_3}(\mathbf{F^{\prime}}^{s_3})+\mathbf{F^{\prime}}^{s_2}_i)+\mathbf{F}^{s_1})+\mathbf{F}
\label{eq8}
\end{equation}
where $f_{s_{1}}$, $f_{s_{2}}$, $f_{s_{3}}$ is the top, middle, bottom GC layer respectively as Fig.~\ref{fig4}, $\mathbf{F}$ is the original DCT coefficients.

Finally, we apply the IDCT to get the human motion frames on Euler angle representation or 3D coordinates.

\subsection{Loss function}
  As in \cite{b24}, when given the Euler angles, we use Mean Angle Error (MAE) as loss function to train our model, and use Mean Per Joint Position Error (MPJPE) proposed in \cite{b29} if given the 3D  coordinates.  Formally, the loss function on the angle and coordinate data can be described by \eqref{eq9} and \eqref{eq10}, respectively:
\begin{equation}
\ell_{a}=\frac{1}{(N+T) K} \sum_{n=1}^{N+T} \sum_{k=1}^{K}\left|\hat{x}_{k, n}-x_{k, n}\right|
\label{eq9}
\end{equation}
\begin{equation}
\ell_{m}=\frac{1}{J(N+T)} \sum_{n=1}^{N+T} \sum_{j=1}^{J}\left\|\hat{\mathbf{p}}_{j, n}-\mathbf{p}_{j, n}\right\|^{2}
\label{eq10}
\end{equation}
where $\hat{x}_{k, n}$ is the predicted $k^{th}$ angles in frame $n$ and $x_{k, n}$ the corresponding ground truth,  $\hat{\mathbf{p}}_{j, n} \in \mathbb{R}^{3}$ denotes the predicted $j^{th}$ joint position at frame $n$, $\mathbf{p}_{j, n}$  is the corresponding ground truth, and $J$ is the number of joints on the human skeleton graph.

\begin{table*}[htb]
	\caption{Short-term prediction in MAE on Human3.6M for the main actions.}	
	\resizebox{\textwidth}{!}{	 ﻿
		\begin{tabular}{lcccccccccccccccccccccccc}
			\toprule
			& \multicolumn{4}{c}{Walking} & \multicolumn{4}{c}{Eating} & \multicolumn{4}{c}{Smoking} & \multicolumn{4}{c}{Directions} & \multicolumn{4}{c}{Greeting} & \multicolumn{4}{c}{Average} \\
			\cmidrule(lr){2-5} \cmidrule(lr){6-9} \cmidrule(lr){10-13} \cmidrule(lr){14-17} \cmidrule(lr){18-21} \cmidrule(lr){22-25}
			milliseconds & 80 & 160 & 320 & 400 & 80 & 160 & 320 & 400 & 80 & 160 & 320 & 400 & 80 & 160 & 320 & 400 & 80 & 160 & 320 & 400 & 80 & 160 & 320 & 400 \\
			\midrule
			DMGNN\cite{b28} & 0.18 & 0.31 & 0.49 & 0.58 & 0.17 & 0.30 & 0.49 & 0.59 & 0.21 & 0.39 & \textbf{0.81} & 0.77 & 0.25 & 0.44 & 0.65 & 0.71 & 0.36 & 0.61 & 0.94 & 1.12 & 0.27 & 0.52 & 0.83 & 0.95 \\
			LTD\cite{b24} & 0.18 & 0.31 & 0.49 & 0.56 & 0.16 & 0.29 & 0.50 & 0.62 & 0.22 & 0.41 & 0.86 & 0.80 & 0.26 & 0.45 & 0.71 & 0.79 & 0.36 & 0.60 & 0.95 & 1.13 & 0.27 & 0.51 & 0.83 & 0.95 \\
			LPJP\cite{b26} & \textbf{0.17} & 0.30 & 0.51 & 0.55 & 0.16 & 0.29 & 0.50 & 0.61 & 0.21 & 0.40 & 0.85 & 0.78 & \textbf{0.22} & \textbf{0.39} & \textbf{0.62} & \textbf{0.69} & 0.34 & 0.58 & 0.94 & 1.12 & \textbf{0.25} & 0.49 & 0.83 & 0.94 \\
			\midrule
			ours & 0.18 & \textbf{0.30} & \textbf{0.47} & \textbf{0.51} & \textbf{0.16} & \textbf{0.28} & \textbf{0.46} & \textbf{0.57} & \textbf{0.21} & \textbf{0.38} & 0.82 & \textbf{0.76} & 0.23 & 0.40 & 0.73 & 0.80 & \textbf{0.34} & \textbf{0.57} & \textbf{0.91} & \textbf{1.08} & 0.26 & \textbf{0.49} & \textbf{0.81} & \textbf{0.92}\\
			\bottomrule
	\end{tabular}}
	\label{tab1}
\end{table*}

\begin{table*}[htb]
	\caption{Short-term prediction in MPJPE on Human3.6M for the main actions.}
	\resizebox{\textwidth}{!}{	
		\begin{tabular}{lcccccccccccccccccccccccc}
			\toprule
			& \multicolumn{4}{c}{Walking} & \multicolumn{4}{c}{Eating} & \multicolumn{4}{c}{Smoking} & \multicolumn{4}{c}{Directions} & \multicolumn{4}{c}{Greeting} & \multicolumn{4}{c}{Average} \\
			\cmidrule(lr){2-5} \cmidrule(lr){6-9} \cmidrule(lr){10-13} \cmidrule(lr){14-17} \cmidrule(lr){18-21} \cmidrule(lr){22-25}
			milliseconds & 80 & 160 & 320 & 400 & 80 & 160 & 320 & 400 & 80 & 160 & 320 & 400 & 80 & 160 & 320 & 400 & 80 & 160 & 320 & 400 & 80 & 160 & 320 & 400 \\
			\midrule
			DMGNN\cite{b28} & 12.5 & 22.9 & 40.1 & 52.5 & 11.0 & 23.2 & 46.1 & 55.9 & 8.0 & 15.4 & 26.4 & 30.2 & 13.0 & 22.4 & \textbf{47.0} & \textbf{57.5} & 17.5 & 33.4 & 73.0 & 89.3 & 14.3 & 29.0 & 58.1 & 70.6 \\
			LTD\cite{b24} & 11.1 & 19.0 & 32.0 & 39.1 & 9.2 & 19.5 & 40.3 & 48.9 & 9.2 & 16.6 & 26.1 & 29.0 & 11.2 & 23.2 & 52.7 & 64.1 & 14.2 & 27.7 & 67.1 & 82.9 & 13.5 & 27.0 & 54.0 & 65.0 \\
			LTD 3D\cite{b24} & 8.9 & 15.7 & 29.2 & 33.4 & 8.8 & 18.9 & 39.4 & 47.2 & 7.8 & 14.9 & 25.3 & 28.7 & 12.6 & 24.4 & 48.2 & 58.4 & 14.5 & 30.5 & 74.2 & 89.0 & 12.1 & 25.0 & 51.0 & 61.3 \\
			LPJP\cite{b26} & 9.6 & 18.0 & 33.1 & 39.1 & 9.1 & 19.5 & 40.2 & 48.8 & 7.2 & 14.2 & 24.7 & 29.7 & \textbf{9.3} & 22.0 & 51.6 & 63.2 & 15.4 & 30.7 & 71.8 & 82.8 & 11.9 & 26.1 & 53.2 & 64.5 \\
			LPJP 3D\cite{b26} & \textbf{7.9} & \textbf{14.5} & 29.1 & 34.5 & 8.4 & \textbf{18.1} & \textbf{37.4} & 45.3 & 6.8 & 13.2 & 24.1 & 27.5 & 11.1 & 22.7 & 48.0 & 58.4 & 13.2 & 28.0 & 64.5 & 77.9 & \textbf{10.7} & 23.8 & 50.0 & 60.2 \\
			\midrule
			ours & 10.2 & 18.0 & 30.5 & 36.6 & 8.9 & 18.7 & 37.5 & 46.0 & 7.5 & 13.6 & 23.1 & 27.4 & 9.5 & \textbf{20.1} & 48.5 & 58.3 & 14.4 & 28.5 & \textbf{62.7} & \textbf{77.6} & 12.5 & 25.5 & 50.8 & 61.8 \\
			ours 3D & 8.1 & 15.0 & \textbf{27.1} & \textbf{31.3} & \textbf{8.2} & 18.4 & 37.7 & \textbf{44.5} & \textbf{6.8} & \textbf{12.9} & \textbf{22.6} & \textbf{27.0} & 10.6 & 22.7 & 48.9 & 60.1 & \textbf{12.8} & \textbf{25.8} & 68.4 & 86.8 & 10.8 & \textbf{23.2} & \textbf{49.3} & \textbf{60.1} \\
			\bottomrule
		\end{tabular}
	}	
	\label{tab2}
\end{table*}

\begin{table}[htp]
	\caption{Long-term prediction in MPJPE on Human3.6M.}
	\centering
	\resizebox{8cm}{!}{
		\begin{tabular}{@{}lcccccccccc@{}}
			\toprule
			& \multicolumn{2}{c}{Walking} & \multicolumn{2}{c}{Eating} & \multicolumn{2}{c}{Smoking} & \multicolumn{2}{c}{Discussion} & \multicolumn{2}{c}{Average} \\ \cmidrule(lr){2-3} \cmidrule(lr){4-5}\cmidrule(lr){6-7} \cmidrule(lr){8-9} \cmidrule(lr){10-11}  
			milliseconds & 560 & 1000 & 560 & 1000 & 560 & 1000 & 560 & 1000 & 560 & 1000 \\ \midrule
			DMGNN\cite{b28} & 56.5 & 90.1 & 80.6 & 107.7 & 39.6 & 61.2 & 96.4 & 111.2 & 68.3 & 92.6 \\
			LTD\cite{b24} & 55.0 & 60.8 & 68.1 & 79.5 & 42.2 & 70.6 & 93.8 & 119.7 & 64.8 & 82.6 \\
			LTD 3D\cite{b24} & 42.3 & 51.3 & 56.5 & 68.6 & 32.3 & 60.5 & 70.5 & 103.5 & 50.4 & 71.0 \\
			LPJP\cite{b26} & 51.8 & 58.7 & 59.3 & 76.5 & 40.3 & 76.8 & 82.6 & 107.7 & 58.5 & 79.9 \\
			LPJP 3D\cite{b26} & \textbf{36.8} & \textbf{41.2} & 58.4 & \textbf{67.9} & 29.2 & 58.3 & 74.0 & 103.1 & 49.6 & 67.6 \\ \midrule
			ours & 40.1 & 45.8 & 60.1 & 74.4 & 30.6 & 59.4 & 71.1 & 83.0 & 50.5 & 65.7 \\
			ours 3D & 37.7 & 43.7 & \textbf{53.0} & 68.7 & \textbf{28.0} & \textbf{55.2} & \textbf{55.7} & \textbf{72.9} & \textbf{43.6} & \textbf{60.1} \\ \bottomrule
	\end{tabular}}
	\label{tab3}
\end{table}

\begin{table*}[htb]
	\caption{Short and long-term prediction in MPJPE on CMU-Mocap dataset.}
	\centering
			\begin{tabular}{@{}lccccccccccccccc@{}}
				\toprule
				& \multicolumn{5}{c}{Basketball} & \multicolumn{5}{c}{Basketball Signal} & \multicolumn{5}{c}{Directing Traffic} \\
				\cmidrule(lr){2-6}\cmidrule(lr){7-11}\cmidrule(lr){12-16}
				milliseconds & 80 & 160 & 320 & 400 & 1000 & 80 & 160 & 320 & 400 & 1000 & 80 & 160 & 320 & 400 & 1000 \\
				DMGNN\cite{b28} & 16.6 & 32.4 & 71.9 & 93.0 & 118.9 & 4.0 & 8.0 & 16.9 & 21.7 & 95.8 & 9.8 & 19.4 & 42.3 & 54.4 & 165.2 \\
				LTD 3D\cite{b24} & 14.0 & 25.4 & 49.6 & 61.4 & 106.1 & 3.5 & 6.1 & 11.7 & 15.2 & 53.9 & 7.4 & 15.1 & 31.7 & 42.2 & 152.4 \\
				LPJP 3D\cite{b26} & 11.6 & 21.7 & 44.4 & 57.3 & \textbf{90.9} & 2.6 & 4.9 & 12.7 & 18.7 & 75.8 & 6.2 & 12.7 & 29.1 & 39.6 & 149.1 \\
				\midrule
				ours & 12.0 & 23.7 & 54.7 & 72.9 & 128.0 & 2.6 & 5.9 & 15.8 & 21.4 & 82.9 & 6.0 & 12.0 & 27.6 & 38.0 & 152.2 \\
				ours 3D & \textbf{10.8} & \textbf{18.9} & \textbf{38.2} & \textbf{49.1} & 97.3 & \textbf{2.2} & \textbf{4.0} & \textbf{10.6} & \textbf{14.8} & \textbf{53.5} & \textbf{5.9} & \textbf{11.5} & \textbf{25.6} & \textbf{34.0} & \textbf{132.8} \\
				\toprule
				& \multicolumn{5}{c}{Jumping} & \multicolumn{5}{c}{Running} & \multicolumn{5}{c}{Soccer} \\
				\cmidrule(lr){2-6}\cmidrule(lr){7-11}\cmidrule(lr){12-16}
				milliseconds & 80 & 160 & 320 & 400 & 1000 & 80 & 160 & 320 & 400 & 1000 & 80 & 160 & 320 & 400 & 1000 \\
				DMGNN\cite{b28} & 21.4 & 44.3 & 96.0 & 118.7 & 191.2 & 11.3 & 17.87 & 21.7 & 26.3 & 69.5 & 15.4 & 32.3 & 68.0 & 80.7 & 167.9 \\
				LTD 3D\cite{b24} & 16.9 & 34.4 & 76.3 & 96.8 & 164.6 & 25.5 & 36.7 & 39.3 & 39.9 & 58.2 & 11.3 & 21.5 & 44.2 & 55.8 & 117.5 \\
				LPJP 3D\cite{b26} & \textbf{12.9} & \textbf{27.6} & \textbf{73.5} & \textbf{92.2} & 176.6 & 23.5 & 34.2 & 35.2 & 36.1 & \textbf{43.1} & 9.2 & 18.4 & 39.2 & \textbf{49.5} & \textbf{93.9} \\
				\midrule
				ours & 13.9 & 29.8 & 77.9 & 102.4 & 177.7 & 21.2 & 29.6 & 27.2 & 28.7 & 75.7 & 9.1 & 18.2 & 42.6 & 53.3 & 125.2 \\
				ours 3D & 13.4 & 29.5 & 74.0 & 96.9 & \textbf{162.1} & \textbf{17.4} & \textbf{21.2} & \textbf{20.6} & \textbf{26.5} & 65.1 & \textbf{9.1} & \textbf{16.7} & \textbf{37.5} & 52.5 & 119.5 \\
				\toprule
				& \multicolumn{5}{c}{Walking} & \multicolumn{5}{c}{Washwindow} & \multicolumn{5}{c}{Average} \\
				\cmidrule(lr){2-6}\cmidrule(lr){7-11}\cmidrule(lr){12-16}
				milliseconds & 80 & 160 & 320 & 400 & 1000 & 80 & 160 & 320 & 400 & 1000 & 80 & 160 & 320 & 400 & 1000 \\
				DMGNN\cite{b28} & 6.8 & 10.8 & 20.0 & 23.8 & 40.1 & 6.1 & 12.8 & 31.5 & 39.7 & 93.3 & 11.4 & 22.2 & 46.0 & 57.2 & 117.7 \\
				LTD 3D\cite{b24} & 7.7 & 11.8 & 19.4 & 23.1 & 40.2 & 5.9 & 11.9 & 30.3 & 40.0 & 79.3 & 11.5 & 20.4 & 37.8 & 46.8 & 96.5 \\
				LPJP 3D\cite{b26} & 6.7 & 10.7 & 21.7 & 27.5 & 37.4 & 5.4 & 11.3 & 29.2 & 39.6 & 79.1 & 9.8 & 17.6 & 35.7 & 45.1 & 93.2 \\
				\midrule
				ours & 7.5 & 12.5 & 19.0 & 23.5 & 67.6 & 4.5 & 10.2 & 31.9 & 44.0 & 90.1 & 9.6 & 17.7 & 37.1 & 48.0 & 112.4 \\
				ours 3D & \textbf{6.3} & \textbf{10.2} & \textbf{17.6} & \textbf{20.5} & \textbf{34.9} & \textbf{4.4} & \textbf{9.6} & \textbf{27.4} & \textbf{37.2} & \textbf{74.9} & \textbf{8.7} & \textbf{15.2} & \textbf{31.4} & \textbf{41.4} & \textbf{92.5} \\ \bottomrule
			\end{tabular}
	\label{tab4}
\end{table*}

\begin{table}[htp]
	\caption{effect of the multiscale designs by MPJPE on Human3.6M.}
	\centering
		\begin{tabular}{@{}lcccc@{}}
			\toprule
			& \multicolumn{4}{c}{Average} \\
			\cmidrule(lr){2-5}
			milliseconds & 80 & 160 & 320 & 400 \\	
			w/o STM & 13.1 & 26.8 & 52.9 & 63.5 \\
			w/o CS-B & 12.6 & 25.8 & 51.9 & 63.1 \\
			w/o coarse-to-fine decoder & 12.9 & 25.9 & 52.1 & 63.1 \\
			\midrule
			with all above & \textbf{12.5} & \textbf{25.5} & \textbf{50.8} & \textbf{61.8}\\
			\bottomrule
		\end{tabular}
	\label{tab5}
\end{table}

\section{Experiments}
In this section, we introduce implementation details, followed by the datasets, the experimental results analysis and ablation study.

\subsection{Implementation details}

The GCN is the cascade of 6 residual blocks, each of which comprises 2 graph convolutional layers. We train the model for 100 epochs  with a learning-rate decay of 0.96 every 2 epochs. The batch size on Human 3.6M dataset is 256 and On CMU dataset is 16. The stack number $N$ of SIM is 3. The feature dim of GCN is set to 256. The feature dim of MLPs in the STM is 16 while that in the CS-B is 512. More implementation details can be found in our project home page at \url{https://github.com/zhouhongh/MGCN}.

\subsection{Datasets}
\subsubsection{Human3.6M}
There are 15 actions performed by 7 subjects for training and testing in the dataset. The actors are represented by a skeleton of 32 joints. Following the settings in \cite{b20,b30}, we remove the global rotations and translations as well as constant angles and down sample the sequences to 25 frames per second.

\subsubsection{CMU-Mocap}
We select 8 actions and report results on the CMU mocap dataset (CMU-Mocap) following \cite{b22,b24,b26,b28}. It is also down sampled to 25 frames per second and removed the global rotations and translations as well as constant angles.

\subsection{Baselines}
We use 3 methods as the baselines: DMGNN \cite{b28}, LTD \cite{b24} and LPJP \cite{b26}. The DMGNN applies the multiscale graph with the ST-GCN backbone. However, it does not adopt the timing modeling method of DCT, and does not use a strategy similar to STM and coarce-to-fine decoding . The LTD proposes the "DCT + space modeling" framework which uses the DCT to encode the trajecory of body joints and apply the fully-connected GCN to capture the huamn body correlations. But it completely ignores the multi-scale idea. The LPJP follows LTD's "DCT + spatial modeling" framework, and uses the popular Transformer \cite{b37} to replace the fully connected GCN and uses a central-to-peripheral ring decoding strategy to predict the human body, which makes LPJP the SOTA method.

Notied that the authors of DMGNN do not report their performance under the MPJPE metric and do not provide the model weight files, we train the DMGNN by ourselves following the settings in their paper and calculate the MPJPE. It is important to note that DMGNN does not contain codes to train directly on 3D coordinate data, we can only train on angle data and then calculate MPJPE. But this does not affect the comparison, because we also provide the MPJPE that calculated from our model trained on the angle data.

\subsection{Results}

For fair comparison, we report both short-term and long-term predictions  for the two datasets, given the the history sequence including 10 frames as input. Noticed that the sequences have the speed of 25 frames per second, The short-term prediction means predicting for 400 milliseconds, 10 frames, and the long-term 1000 miliseconds, 25 frames.

\subsubsection{Results on Human 3.6M}

For short-term prediction, we evaluate our method under both MAE (Table~\ref{tab1}) and MPJPE (Table~\ref{tab2}) protocols, in comparison to state-of-the-art baselines\cite{b22,b24,b26,b28}. The 3D suffix to a method indicates the method is directly trained on 3D joint positions. Otherwise, the results were obtained by converting the joints angles to 3D positions.
Due to the limited space, we report the results of the five main actions and the average results under all 15 actions. On the average results, we can see that our approach outperforms all the baselines at the later time steps , but a little worse than the SOTA method \cite{b26} at 80 milliseconds, which indicates that our approach mainly works at farther time steps. we speculate that it is because that at the closer time steps the movement is subtle and do not need too complicated spatial encoding. This problem also can be seen on walking and direction in Table~\ref{tab2}. In particular, let's look at the performance of DMGNN, which also uses a multi-scale strategy. Although DMGNN slightly outperformed other methods including ours method for a few moments, such as 320ms for Smoking in Table~\ref{tab1}, 320ms for Drections, and 400ms for Drections in Table~\ref{tab2}, our method leads the way in most categories, especially in the average metrics. It is proved that our method has significant advantages over DMGNN's multi-scale modeling methods, and even slightly superior to the current best method LPJP \cite{b26}.

Additionally, our method is superior to LTD \cite{b24} in all categories in Table~\ref{tab1}, \ref{tab2} and \ref{tab3},  which shows that the single scale space modeling strategy is insufficient to capture the complicated human body correlations and it is necessary to introduce the multi-scale strategy in the "DCT + space modeling" framework. It needs to be pointed out that, as indicated in \cite{b24}, the MAE metric may incorrectly evaluate the results on account of the cyclicity of angles, so we only report the results under MPJPE in the following experiments (long-term experiment on Human3.6M, short-term and long-term experiments on CMU and ablation study). 

We also compare our results with baselines \cite{b28,b24,b26} in long-term scenarios in Table~\ref{tab3}. On the average metric, our method achieves a greater improvement than short-term scenarios and outperformes all baselines, indicating that our method can still maintain good accuracy in long-term prediction, thanks to our multi-scale strategy which fully captured human body correlations. However, in some actions, such as walking and eating, our method is slightly inferior to the SOTA method. We note that both walking and eating are strongly periodic movements, and such movements are easy to capture their body correlations. Therefore, the single-scale approach may be sufficient, while the improvement brought by the multi-scale strategy is not significant in these categories.

\subsubsection{Results on CMU-Mocap dataset}

To verify the universality of our approach, we also reported the results on CMU-Mocap dataset in both short and long-term scenarios by MPJPE as in Table~\ref{tab4}. First, let's focus on the average indicator that best reflects the overall performance of the model. Looking at the methods with the 3D suffix, our method outperforms all baselines on average, which proves that the advantage of our method is stable and still has excellent performance on other data sets. And the results for methods without 3D suffix show that although both DMGNN and our approach adopt a multi-scale strategy, it is clear that our approach outperforms DMGNN on most categories, demonstrating that our multi-scale approach makes better use of the body associations of the human body.

In the long-term prediction of running, basketball, soccer, our method is somewhat inferior to the SOTA method, which may be because these categories all contain large periodic movements such as running, and can be well modeled without the need for multi-scale strategies. However, success in predicting more categories of actions and moments of human motions still shows significant advantages of our method.

\subsection{Ablation study}

We quantify the effect of our designs for the multiscale graph, including  STM, CS-B and the coarse-to-fine decoder.
The STM generates the larger graphs by adaptively aggregating the body joints to components. The CS-B exchange information among different scales, and the coarse-to-fine decoder decode the human motions sequentially. Noticed that directly remove the STM will cause the collapse of the multiscale architecture, we replace it with the average strategy in \cite{b28} which simply forms the multiscale graphs by computing the mean value of body joints. Similarly, we replace the coarse-to-fine decoder by the simply  parallel strategy which directly summing the outputs of three scales.

We train the model with all the 15 actions, and show the average value in Table~\ref{tab5}. We can see that the prediction errors increase no matter we remove any of the three parts. And the STM brings the biggest improvement, because it avoided the drop of information compared the average strategy in \cite{b28}.

\section{Conclusion and future work}

Human motion prediction has gained more and more attention with the rapid development of human-robot interaction and  autonomous driving.  Capturing of the human body correlations is the key to predict future motions. We propose the MGCN to explore the correlations by the multiscale graphs in and corss scales and our  exhaustive experiments demonstrate that the proposed method outperform the state-of-the-arts methods especially for those complicated and aperiodic actions.

Further more, we note that current studies have focused on motion prediction for single person, with little consideration for multi-person scenarios. However, in real life, most human motions involve interactions with others, so it is obviously not sufficient to model a person without such interaction information. The next step of our work is to mine the interaction information among multiple persons, so as to achieve more accurate motion prediction in multi-person scenarios.

\end{document}